\documentclass{article}
\usepackage{cite}
\usepackage{amsmath,amssymb,amsfonts,amsthm}
\usepackage{bm}
\usepackage{algorithmic}
\usepackage{graphicx}
\usepackage{textcomp}
\usepackage{xcolor}
\usepackage{bbm}
\usepackage{dsfont}
\usepackage{subcaption}

\usepackage[font=small]{caption}

\usepackage{booktabs,makecell,multirow,hhline,threeparttable} 
\usepackage{xcolor,colortbl,pgf}
\definecolor{Color1}{RGB}{240, 240, 240}
\usepackage{makecell}

\usepackage{caption}
\usepackage{graphicx} %
\usepackage[hyphens]{url}  %
\usepackage{enumitem}

\usepackage{xspace}
\usepackage{multirow}
\usepackage{booktabs}
\usepackage[pagebackref,breaklinks,colorlinks=true,citecolor=blue]{hyperref}
\usepackage{cleveref}

\newcommand{\fix}[1]{\textcolor{blue}{#1}}

\renewcommand{\paragraph}[1]{\noindent\textbf{#1}}

\def\BibTeX{{\rm B\kern-.05em{\sc i\kern-.025em b}\kern-.08em
    T\kern-.1667em\lower.7ex\hbox{E}\kern-.125emX}}

\newcounter{packednmbr}

\DeclareUnicodeCharacter{2212}{-}

\usepackage{amsmath,amsfonts,bm}

\def\eqref#1{equation~\ref{#1}}

\def\1{\bm{1}}

\def\vx{{\bm{x}}}

\DeclareMathAlphabet{\mathsfit}{\encodingdefault}{\sfdefault}{m}{sl}
\SetMathAlphabet{\mathsfit}{bold}{\encodingdefault}{\sfdefault}{bx}{n}

\usepackage[ruled,vlined]{algorithm2e}

\SetCommentSty{mycommfont}
\usepackage{pifont}
\newcommand{\cmark}{\ding{51}}%
\newcommand{\xmark}{\ding{55}}%

\newcommand{\method}{Sparse Continual Learning\xspace}
\newcommand{\methodabbr}{SparCL\xspace}

\newcommand{\sparseweight}{task-aware dynamic masking\xspace}
\newcommand{\sparseweightabbr}{TDM\xspace}
\newcommand{\sparsedata}{dynamic data removal\xspace}
\newcommand{\sparsedataabbr}{DDR\xspace}
\newcommand{\sparsegrad}{dynamic gradient masking\xspace}
\newcommand{\sparsegradabbr}{DGM\xspace}

\usepackage[final]{neurips_2022}

\usepackage[utf8]{inputenc} 
\usepackage[T1]{fontenc}    
\usepackage{hyperref}       
\usepackage{url}            
\usepackage{booktabs}       
\usepackage{amsfonts}       
\usepackage{nicefrac}       
\usepackage{microtype}      
\usepackage{xcolor}         
\usepackage{gensymb}
\setcitestyle{numbers}

\title{SparCL: Sparse Continual Learning on the Edge}

\author{%
Zifeng Wang\text{$^{1,\dagger}$},
  Zheng Zhan\text{$^{1,\dagger}$}, 
 Yifan Gong$^{1}$, Geng Yuan$^{1}$, Wei Niu$^{2}$, Tong Jian$^{1}$,\\ \textbf{Bin Ren$^{2}$, Stratis Ioannidis$^{1}$, Yanzhi Wang$^{1}$, Jennifer Dy$^{1}$}
 \\
  $^{1}$ Northeastern University, $^{2}$ College of William and Mary
\\
  \texttt{\{zhan.zhe, gong.yifa, geng.yuan, yanz.wang\}@northeastern.edu}, \\
 \texttt{\{zifengwang, jian, ioannidis, jdy\}@ece.neu.edu}, \\
  \texttt{wniu@email.wm.edu, bren@cs.wm.edu}}

\begin{document}

\maketitle
\def\thefootnote{$\dagger$}\footnotetext{Both authors contributed equally to this work}

\begin{abstract}
Existing work in continual learning (CL) focuses on mitigating catastrophic forgetting, \emph{i.e.}, model performance deterioration on past tasks when learning a new task. However, the training efficiency of a CL system is under-investigated, which limits the real-world application of CL systems under resource-limited scenarios. In this work, we propose a novel framework called \method (\methodabbr), 
which is the first study that leverages sparsity to enable cost-effective continual learning on edge devices. \methodabbr achieves both training acceleration and accuracy preservation through the synergy of three aspects: \emph{weight sparsity}, \emph{data efficiency}, and \emph{gradient sparsity}. Specifically, we propose task-aware dynamic masking (TDM) to learn a sparse network throughout the entire CL process, dynamic data removal (DDR) to remove less informative training data, and dynamic gradient masking (DGM) to sparsify the gradient updates. Each of them not only improves efficiency, but also further mitigates catastrophic forgetting.  \methodabbr consistently improves the training efficiency of existing state-of-the-art (SOTA) CL methods by at most $23\times$ less training FLOPs, and, surprisingly, further improves the SOTA accuracy by at most $1.7\%$. \methodabbr also outperforms competitive baselines obtained from adapting SOTA sparse training methods to the CL setting in both efficiency and accuracy. We also evaluate the effectiveness of \methodabbr on a real mobile phone, further indicating the practical potential of our method. Source code will be released.

\end{abstract}

\section{Introduction} \label{sec:intro}

The objective of Continual Learning (CL) is to enable an intelligent system to accumulate knowledge from a sequence of tasks, such that it exhibits satisfying performance on both old and new tasks~\cite{kirkpatrick2017overcoming}. Recent methods mostly focus on addressing the \emph{catastrophic forgetting}~\cite{mccloskey1989catastrophic} problem -- learning model tends to suffer performance deterioration on previously seen tasks. However, in the real world, when the CL applications are deployed in resource-limited platforms~\cite{pellegrini2021continual} such as edge devices, the learning efficiency, w.r.t. both training speed and memory footprint, are also crucial metrics of interest, yet they are rarely explored in prior CL works.

Existing CL methods can be categorized into regularization-based~\cite{kirkpatrick2017overcoming,zenke2017continual,li2017learning,aljundi2018memory}, rehearsal-based~\cite{rebuffi2017icarl, wu2019large, chaudhry2020using, buzzega2020dark}, and architecture-based~\cite{mallya2018packnet, serra2018overcoming, ke2020continual, wang2021learning,wang2022dualprompt,zhao2022deep}. Both regularization- and rehearsal-based methods directly train a dense model, which might even be over-parametrized for the union of all tasks~\cite{dong2019network, ma2020image}; Though several architecture-based methods~\cite{rusu2016progressive, wang2020learn, yang2021grown} start with a sparse sub-network from the dense model, they still grow the model size progressively to learn emerging tasks. The aforementioned methods, although striving for greater performance with less forgetting, still introduce significant memory and computation overhead during the whole CL process.

\begin{figure*} [t]
     \centering
     \includegraphics[width=0.78\textwidth]{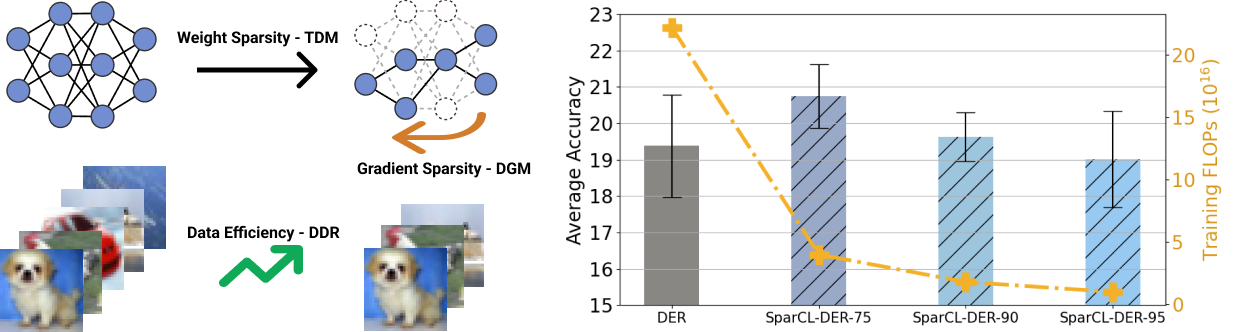}  
     \caption{\textbf{Left:} Overview of \methodabbr. \methodabbr consists of three complementary components: \sparseweight (\sparseweightabbr) for weight sparsity, \sparsedata (\sparsedataabbr) for data efficiency, and \sparsegrad (\sparsegradabbr) for gradient sparsity. \textbf{Right:} \methodabbr successfully preserves the accuracy and significantly improves efficiency over DER++~\cite{buzzega2020dark}, one of the SOTA CL methods, with different sparsity ratios on the Split Tiny-ImageNet~\cite{deng2009imagenet} dataset.}
    \label{fig:intro} 
    \vspace{-0.1in}
\end{figure*}

Recently, another stream of work, sparse training~\cite{lee2018snip,evci2020rigging,bellec2018deep} has emerged as a new training trend to achieve training acceleration, which embraces the promising training-on-the-edge paradigm. With sparse training, each iteration takes less time with the reduction in computation achieved by sparsity, under the traditional i.i.d.~learning setting.
Inspired by these sparse training methods, we naturally think about introducing sparse training to the field of CL.
A straightforward idea is to directly combine existing sparse training methods, such as SNIP~\cite{lee2018snip}, RigL~\cite{evci2020rigging}, with a rehearsal buffer under the CL setting. However, these methods fail to consider key challenges in CL to mitigate catastrophic forgetting, for example, properly handling transition between tasks.
As a result, these sparse training methods, though enhancing training efficiency, cause significant accuracy drop (see Section~\ref{sec:main_results}). Thus, we would like to explore a general strategy, which is orthogonal to existing CL methods, that not only leverages the idea of sparse training for efficiency, but also addresses key challenges in CL to preserve (or even improve) accuracy.


In this work, we propose \emph{\method} (\methodabbr), a general framework for cost-effective continual learning, aiming at enabling practical CL on edge devices. As shown in Figure~\ref{fig:intro} (left), \methodabbr achieves both learning acceleration and accuracy preservation through the synergy of three aspects: \emph{weight sparsity}, \emph{data efficiency}, and \emph{gradient sparsity}. Specifically, to maintain a small dynamic sparse network during the whole CL process, we develop a novel task-aware dynamic masking (\sparseweightabbr) strategy to keep only important weights for both the current and past tasks, with special consideration during task transitions. 
Moreover, we propose a dynamic data removal (\sparsedataabbr) scheme, which progressively removes ``easy-to-learn'' examples from training iterations, which further accelerates the training process and also improves accuracy of CL by balancing current and past data and keeping more informative samples in the buffer. Finally, we provide an additional dynamic gradient masking (\sparsegradabbr) strategy to leverage gradient sparsity for even better efficiency and knowledge preservation of learned tasks, such that only a subset of sparse weights are updated. Figure~\ref{fig:intro} (right) demonstrates that \methodabbr successfully preserves the accuracy and significantly improves efficiency over DER++~\cite{buzzega2020dark}, one of the SOTA CL methods, under different sparsity ratios.

\methodabbr is simple in concept, compatible with various existing rehearsal-based CL methods, and efficient under practical scenarios. We conduct comprehensive experiments on multiple CL benchmarks to evaluate the effectiveness of our method. We show that \methodabbr works collaboratively with existing CL methods, greatly accelerates the learning process under different sparsity ratios, and even sometimes improves upon the state-of-the-art accuracy. We also establish competing baselines by combining representative sparse training methods with advanced rehearsal-based CL methods. \methodabbr again outperforms these baselines in terms of both efficiency and accuracy. Most importantly, we evaluate our \methodabbr framework on real edge devices to demonstrate the practical potential of our method. We are not aware of any prior CL works that explored this area and considered the constraints of limited resources during training.

In summary, our work makes the following contributions:
\begin{itemize}[leftmargin=*]
   \item We propose \emph{\method} (\methodabbr), a general framework for cost-effective continual learning, which achieves learning acceleration through the synergy of \emph{weight sparsity}, \emph{data efficiency}, and \emph{gradient sparsity}. To the best of our knowledge, our work is the first to introduce the idea of sparse training to enable efficient CL on edge devices.
    \item \methodabbr shows superior performance compared to both conventional CL methods and CL-adapted sparse training methods on all benchmark datasets, leading to at most $23\times$ less training FLOPs and, surprisingly, $1.7\%$ improvement over SOTA accuracy.
    \item We evaluate \methodabbr on a real mobile edge device, demonstrating the practical potential of our method and also encouraging future research on CL on-the-edge. The results indicate that our framework can achieve at most $3.1\times$ training acceleration. 
\end{itemize}
\section{Related work}
\subsection{Continual Learning}
The main focus in continual learning (CL) has been mitigating catastrophic forgetting. Existing methods can be classified into three major categories. \emph{Regularization-based} methods~\cite{kirkpatrick2017overcoming, zenke2017continual, li2017learning, aljundi2018memory} limit updates of important parameters for the prior tasks by adding corresponding regularization terms. While these methods reduce catastrophic forgetting to some extent, their performance deteriorates under challenging settings~\cite{mai2021online}, and on more complex benchmarks~\cite{rebuffi2017icarl, wu2019large}. \emph{Rehearsal-based} methods~\cite{chaudhry2018efficient, chaudhry2019tiny, hayes2019memory} save examples from previous tasks into a small-sized buffer to train the model jointly with the current task. Though simple in concept, the idea of rehearsal is very effective in practice and has been adopted by many state-of-the-art methods~\cite{buzzega2020dark, cha2021co2l, pham2021dualnet}. \emph{Architecture-based} methods~\cite{rusu2016progressive, mallya2018packnet, wang2020learn, wang2021learning, yan2021dynamically} isolate existing model parameters or assign additional parameters for each task to reduce interference among tasks. 
As mentioned in Section~\ref{sec:intro}, most of these methods use a dense model without consideration of efficiency and memory footprint, thus \fix{are} not applicable to resource-limited settings. Our work, orthogonal to these methods, serves as a general framework for making these existing methods efficient and enabling a broader deployment, \emph{e.g.}, CL on edge devices.


A limited number of works explore sparsity in CL, however, for different purposes. Several methods \cite{wang2020learn, mallya2018packnet, mallya2018piggyback, sokar2021spacenet} incorporate the idea of weight pruning~\cite{han2015learning} to allocate a sparse sub-network for each task to reduce inter-task interference. Nevertheless, these methods still reduce the full model sparsity progressively for every task and finally end up with a much denser model. On the contrary, \methodabbr maintains a sparse network throughout the whole CL process, introducing great efficiency and memory benefits both during training and at the output model. A recent work~\cite{chen2020long} aims at discovering lottery tickets~\cite{frankle2018lottery} under CL, but still does not address efficiency. However, the existence of lottery tickets in CL serves as a strong justification for the outstanding performance of our~\methodabbr.

\subsection{Sparse Training}

There are two main approaches for sparse training: fixed-mask sparse training and dynamic sparse training. Fixed-mask sparse training methods \cite{lee2018snip,wang2019picking,tanaka2020pruning,wimmer2020freezenet} first apply pruning, then execute traditional training on the sparse model with the obtained fixed mask. The pre-fixed structure limits the accuracy performance, and the first stage still causes huge computation and memory consumption. To overcome these drawbacks, dynamic mask methods \cite{evci2020rigging,dettmers2019sparse,mostafa2019parameter,mocanu2018scalable,bellec2018deep} adjust the sparsity topology during training while maintaining low memory footprint. These methods start with a sparse model structure from an untrained dense model, then combine sparse topology exploration at the given sparsity ratio with the sparse model training. Recent work \cite{yuan2021mest} further considers to incorporate data efficiency into sparse training for better training accelerations. However, all prior sparse training works are focused on the traditional training setting, while CL is a more complicated and difficult scenario with inherent characteristics not explored by these works. In contrast to prior sparse training methods, our work explores a new learning paradigm that introduces sparse training into CL for efficiency and also addresses key challenges in CL, mitigating catastrophic forgetting.

\section{Continual Learning Problem Setup}
In supervised CL, a model $f_\theta$ learns from a sequence of tasks $\mathcal{D} = \{\mathcal{D}_1, \ldots, \mathcal{D}_T\}$, where each task $\mathcal{D}_t = \{(\vx_{t, i}, y_{t, i})\}_{i=1}^{n_t}$ consists of input-label pairs, and each task has a disjoint set of classes. Tasks arrive sequentially, and the model must adapt to them.
At the $t$-th step, the model gains access to data from the $t$-th task. However, a small fix-sized rehearsal buffer $\mathcal{M}$ is allowed to save data from prior tasks. At test time, the easiest setting is to assume task identity is known for each coming test example, named task-incremental learning (Task-IL). If this assumption does not hold, we have the more difficult class-incremental learning (Class-IL) setting. In this work, we mainly focus on the more challenging Class-IL setting, and only report Task-IL performance for reference.

The goal of conventional CL is to train a model sequentially that performs well on all tasks at test time. The main evaluation metric is average test accuracy on all tasks. In real-world resource-limited scenarios, we should further consider \emph{training efficiency} of the model. Thus, we measure the performance of the model more comprehensively by including training FLOPs and memory footprint.

\section{\method (\methodabbr)}
\begin{figure*} [t]
     \centering
     \includegraphics[width=\textwidth]{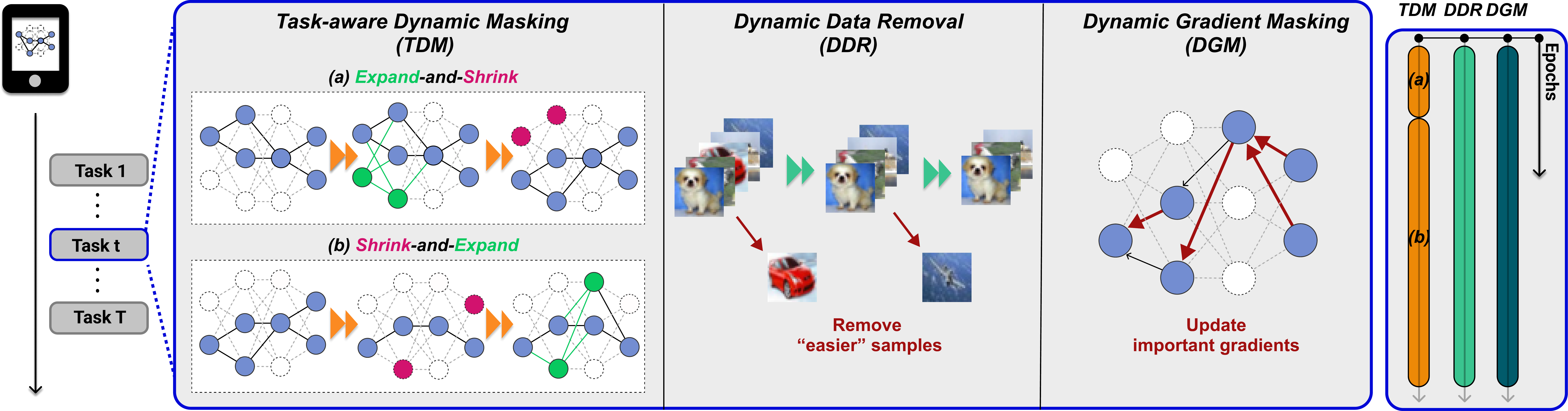}  
     \caption{Illustration of the \methodabbr workflow. Three components work synergistically to improve training efficiency and further mitigate catastrophic forgetting for preserving accuracy.}
    \label{fig:overall_flow}  
\vspace{-0.1in}
\end{figure*}

Our method, \method, is a unified framework composed of three complementary components: \emph{\sparseweight} for weight sparsity, \emph{\sparsedata} for data efficiency, and \emph{\sparsegrad} for gradient sparsity. The entire framework is shown in Figure~\ref{fig:overall_flow}. We will illustrate each component in detail in this section.

\subsection{Task-aware Dynamic Masking} \label{sec:TDM}
To enable cost-effective CL in resource limited scenarios, \methodabbr is designed to maintain a dynamic structure when learning a sequence of tasks, such that it not only achieves high efficiency, but also intelligently adapts to the data stream for better performance. Specifically, we propose a strategy named \emph{\sparseweight} (\sparseweightabbr), which dynamically removes less important weights and grows back unused weights for stronger representation power periodically by maintaining a single binary weight mask throughout the CL process. Different from typical sparse training work, which only leverages the weight magnitude~\cite{mocanu2018scalable} or the gradient w.r.t. data from a single training task~\cite{evci2020rigging, yuan2021mest}, \sparseweightabbr considers also the importance of weights w.r.t. data saved in the rehearsal buffer, as well as the switch between CL tasks.

Specifically, \sparseweightabbr strategy starts from a randomly initialized binary mask $M_{\theta} = M_{0}$, with a given sparsity constraint ${\|M_\theta\|_0}/{\|\theta\|_0} = 1-s$, where $s \in [0, 1]$ is the sparsity ratio. Moreover, it makes different intra- and inter-task adjustments to keep a dynamic sparse set of weights based on their continual weight importance (CWI). We summarize the process of task-aware dynamic masking in Algorithm~\ref{alg:TDM} and elaborate its key components in detail below.

\textbf{Continual weight importance (CWI).} For a model $f_\theta$ parameterized by $\theta$, the CWI of weight $w \subset \theta$ is defined as follows:
\begin{equation} \label{eq:cwi}
    \operatorname{CWI}(w) = \|w\|_{1} + \alpha \|\frac{\partial \Tilde{\mathcal{L}}(\mathcal{D}_t; \theta) }{\partial w}\|_{1} + \beta \|\frac{\partial \mathcal{L}(\mathcal{M}; \theta)}{\partial w}\|_{1}, 
\end{equation}
where $\mathcal{D}_t$ denotes the training data from the $t$-th task, $\mathcal{M}$ is the current rehearsal buffer, and $\alpha$, $\beta$ are coefficients to control the influence of current and buffered data, respectively. Moreover, $\mathcal{L}$ represents the cross-entropy loss for classification, while $\Tilde{\mathcal{L}}$ is the \emph{single-head}~\cite{ahn2021ss} version of the cross-entropy loss, which only considers classes from the current task by masking out the logits of other classes.

Intuitively, CWI ensures we keep (1) weights of larger magnitude for output stability, (2) weights important for the current task for learning capacity, and (3) weights important for past data to mitigate catastrophic forgetting. Moreover, inspired by the classification bias in CL~\cite{ahn2021ss}, we use the single-head cross-entropy loss when calculating importance score w.r.t.\ the current task to make the importance estimation more accurate.

\textbf{Intra-task adjustment.} When training the $t$-th task, a natural assumption is that the data distribution is consistent inside the task, thus we would like to update the sparse model in a relatively stable way while keeping its flexibility. Thus, in Algorithm~\ref{alg:TDM}, we choose to update the sparsity mask $M_\theta$ in a \emph{shrink-and-expand} way every $\delta k$ epochs. We first remove $p_{\texttt{intra}}$ of the weights of least CWI to retain learned knowledge so far. Then we randomly select unused weights to recover the learning capacity for the model and keep the sparsity ratio $s$ unchanged.


\textbf{Inter-task adjustment.} When tasks switches, on the contrary, we assume data distribution shifts immediately. Ideally, we would like the model to keep the knowledge learned from old tasks as much as possible, and to have enough learning capacity to accommodate the new task. Thus, instead of the shrink-and-expand strategy for intra-task adjustment, we follow an \emph{expand-and-shrink} scheme. Specifically, at the beginning of the $(t+1)$-th task, we expand the sparse model by randomly adding a proportion of $p_\texttt{inter}$ unused weights. Intuitively, the additional learning capacity facilitates fast adoption of new knowledge and reduces interference with learned knowledge. We allow our model to have smaller sparsity (\emph{i.e.}, larger learning capacity) temporarily for the first $\delta k$ epochs as a warm-up period, and then remove the $p_\texttt{inter}$ weights with least CWI, following the same process in the intra-task case, to satisfy the sparsity constraint.


\begin{algorithm}[t]

\SetAlgoLined
\SetNoFillComment
\textbf{Input}: Model weight $\theta$, number of tasks $T$, training epochs of the $t$-th task $K_t$, binary sparse mask $M_\theta$, sparsity ratio $s$, intra-task adjustment ratio $p_{\texttt{intra}}$, inter-task adjustment ratio $p_{\texttt{inter}}$, update interval $\delta k$ \\
\textbf{Initialize:} $\theta$, $M_\theta$, s.t. ${\|M_\theta\|_0}/{\|\theta\|_0} = 1-s$ \\

\For{$t = 1,\ldots,T$}{
    \For{$e = 1, \ldots, K_T$}{
    
        \If{$t > 1$}{ 
        \tcc{Inter-task adjustment}
        Expand $M_\theta$ by randomly adding unused weights, \\
        \quad s.t. ${\|M_\theta\|_0}/{\|\theta\|_0} = 1-(s-p_{\texttt{inter}})$\\
        \If{$e = \delta k$}{
        Shrink $M_\theta$ by removing the least important weights according to \eqref{eq:cwi}, \\
        \quad s.t. ${\|M_\theta\|_0}/{\|\theta\|_0} = 1-s$ \\
        }
        }
        \If{$e\ \operatorname{mod}\ \delta k = 0$}{
        \tcc{Intra-task adjustment}
        Shrink $M_\theta$ by removing the least important weights according to \eqref{eq:cwi}, \\
        \quad s.t. ${\|M_\theta\|_0}/{\|\theta\|_0} = 1-(s+p_{\texttt{intra}})$ \\
        Expand $M_\theta$ by randomly adding unused weights, \\ \quad s.t. ${\|M_\theta\|_0}/{\|\theta\|_0} = 1-s$\\
        }
        Update $\theta \odot M_\theta$ via backpropagation \\
    }

 }
 \caption{Task-aware Dynamic Masking~(\sparseweightabbr)}
 \label{alg:TDM}
\end{algorithm}

\subsection{Dynamic Data Removal} \label{sec:DDR}

In addition to weight sparsity, decreasing the amount of training data can be directly translated into the saving of training time without any requirements for hardware support. Thus, we would also like to explore data efficiency to reduce the training workload. 
Some prior CL works select informative examples to construct the rehearsal buffer~\cite{aljundi2019gradient, borsos2020coresets, yoon2021online}. However, the main purpose of them is not training acceleration, thus they either introduce excessive computational cost or consider different problem settings. 
By considering the features of CL, we present a simple yet effective strategy, \emph{\sparsedata} (\sparsedataabbr), to reduce training data for further acceleration.

We measure the importance of each training example by the occurrence of misclassification~\cite{toneva2018empirical, yuan2021mest} during CL. In \sparseweightabbr, the sparse structure of our model updates periodically every $\delta k$ epochs, so we align our data removal process with the update of weight mask for further efficiency and training stability. In Section~\ref{sec:TDM}, we have partitioned the training process for the $t$-th task into $N_t = K_t / \delta k$ stages based on the dynamic mask update. Therefore, we gradually remove training data at the end of $i$-th stage, based on the following policy: 1) Calculate the total number of misclassifications $f_i(x_j)$ for each training example during the $i$-th stage. 2) Remove a proportion of $\rho_i$ training samples with the least number of misclassifications. Although our main purpose is to keep the ``harder'' examples to learn to consolidate the sparse model, we can get further benefits for better CL result. First, the removal of ``easier'' examples increases the probability that ``harder'' examples to be saved to the rehearsal buffer, given the common strategy, \emph{e.g.} reservoir sampling~\cite{chaudhry2019tiny}, to buffer examples. Thus, we construct a more informative buffer in a implicit way without heavy computation. Moreover, since the buffer size is much smaller than the training set size of each task, the data from the buffer and the new task is highly imbalanced, \sparsedata also relieves the data imbalance issue.

Formally, we set the data removal proportion for each task as $\rho \in [0, 1]$, and a cutoff stage, such that:
\begin{equation}
    \sum_{i=1}^{\texttt{cutoff}} \rho_i = \rho, \quad \sum_{i=\texttt{cutoff}+1}^{N_k} \rho_i = 0
\end{equation}
The cutoff stage controls the trade-off between efficiency and accuracy: When we set the cutoff stage earlier, we reduce the training time for all the following stages; however, when the cutoff stage is set too early, the model might underfit the removed training data. Note that when we set $\rho_i = 0$ for all $i = 1, 2, \ldots, N_t$ and $\texttt{cutoff}=N_t$, we simply recover the vanilla setting without any data efficiency considerations. In our experiments, we assume $\rho_i = \rho / \texttt{cutoff}$, i.e., removing equal proportion of data at the end of every stage, for simplicity. We also conduct comprehensive exploration study for $\rho$ and the selection of the cutoff stage in Section~\ref{sec:ablation} and Appendix~\ref{app:cutoff}.



\subsection{Dynamic Gradient Masking} \label{sec:DGM}
With \sparseweightabbr and \sparsedataabbr, we can already achieve bi-level efficiency during training. To further boost training efficiency,  
we explore sparsity in gradient and propose \sparsegrad (\sparsegradabbr) for CL. 
Our method focuses on reducing computational cost by only applying the most important gradients onto the corresponding unpruned model parameters via a gradient mask. The gradient mask is also dynamically updated along with the weight mask defined in Section~\ref{sec:TDM}. Intuitively, while targeting for better training efficiency, \sparsegradabbr also promotes the preservation of past knowledge by preventing a fraction of weights from update.

 
Formally, our goal here is to find a subset of unpruned parameters (or, equivalently, a gradient mask $M_G$) to update over multiple training iterations. For a model $f_\theta$ parameterized by $\theta$, we have the corresponding gradient matrix $G$ calculated during each iteration. To prevent the pruned weights from updating, the weight mask $M_\theta$ will be applied onto the gradient matrix $G$ as $G\odot M_\theta$ during backpropagation. Besides the gradients of pruned weights, we in addition consider to remove less important gradients for faster training. To achieve this, we introduce the continual gradient importance (CGI) based on the CWI to measure the importance of weight gradients.
\begin{equation} \label{eq:cgi}
    \operatorname{CGI}(w) = \alpha \|\frac{\partial \Tilde{\mathcal{L}}(\mathcal{D}_t; \theta) }{\partial w}\|_{1} + \beta \|\frac{\partial \mathcal{L}(\mathcal{M}; \theta)}{\partial w}\|_{1}. 
\end{equation}
We remove a proportion $q$ of non-zero gradients from $G$ with less importance measured by CGI and we have $\|M_G\|_0/\|\theta\|_0 = 1-(s+q)$. The gradient mask $M_G$ is then applied onto the gradient matrix $G$. During the entire training process, the gradient mask $M_G$ is updated with a fixed interval.

\section{Experiment} \label{sec:experiment}

\subsection{Experiment Setting} \label{sec:exp_setting}
\textbf{Datasets.} We evaluate our \methodabbr on two representative CL benchmarks, Split CIFAR-10~\cite{krizhevsky2009learning} and Split Tiny-ImageNet~\cite{deng2009imagenet} to verify the efficacy of \methodabbr. In particular, we follow~\cite{buzzega2020dark,zenke2017continual} by splitting CIFAR-10 and Tiny-ImageNet into 5 and 10 tasks, each of which consists of 2 and 20 classes respectively. Dataset licensing information can be found in Appendix~\ref{app:license}.

\textbf{Comparing methods.} 
In particular, we select CL methods of all kinds including regularization-based (EWC~\cite{kirkpatrick2017overcoming}, LwF~\cite{li2017learning}), architecture-based (PackNet~\cite{mallya2018packnet}, LPS~\cite{wang2020learn}), and rehearsal-based (A-GEM~\cite{chaudhry2018efficient}, iCaRL~\cite{mehtaempirical}, FDR~\cite{benjamin2018measuring}, ER~\cite{chaudhry2019tiny}, DER++~\cite{buzzega2020dark}) methods. Note that PackNet and LPS are only compatible with task-incremental learning. We also adapt representative sparse training methods (SNIP~\cite{lee2018snip}, RigL~\cite{evci2020rigging}) to the CL setting by combining them with DER++ (SNIP-DER++, RigL-DER++).

\textbf{Variants of our method.} To show the generality of \methodabbr, we combine it with DER++ (one of the SOTA CL methods), and ER (simple and widely-used) as \emph{\methodabbr-DER++} and  \emph{\methodabbr-ER}, respectively. We also vary the weight sparsity ratio ($0.75, 0.90, 0.95$) of \methodabbr for a comprehensive evaluation.


\textbf{Evaluation metrics.} We use the average accuracy on all tasks to evaluate the performance of the final model. Moreover, we evaluate the training FLOPs~\cite{evci2020rigging}, and memory footprint~\cite{yuan2021mest} (including feature map pixels and model parameters during training) to demonstrate the efficiency of each method. Please see Appendix~\ref{app:metrics} for detailed definitions of these metrics.


\textbf{Experiment details.} For fair comparison, we strictly follow the settings in prior CL work~\cite{buzzega2020dark, hsu2018re}. We sets the per task training epochs to $50$ and $100$ for Split CIFAR-10 and Tiny-ImageNet, respectively, with a batch size of $32$. For the model architecture, We follow~\cite{rebuffi2017icarl, buzzega2020dark} and adopt the ResNet-18~\cite{he2016deep} without any pre-training. We also use the best hyperparameter setting reported in~\cite{buzzega2020dark, wang2020learn} for CL methods, and in~\cite{lee2018snip, evci2020rigging} for CL-adapted sparse training methods. For \methodabbr and its competing CL-adapted sparse training methods, we adopt a uniform sparsity ratio for all convolutional layers. Please see Appendix~\ref{app:expri_overall} for other details. 

\subsection{Main Results} \label{sec:main_results}
    

\begin{table*}[t!]
\caption{Comparison with CL methods. \methodabbr consistently improves training efficiency of the corresponding CL methods while preserves (or even improves) accuracy on both class- and task-incremental settings.
}
\label{table:cifar_imagenet_CL}
\begin{center}
\scalebox{0.7}{
\begin{tabular}{l||cc|ccc|ccc}
\toprule 
 \multirow{3}{*}{\textbf{Method}}  & \multirow{3}{*}{\textbf{Sparsity}} & \multirow{3}{*}{\textbf{Buffer size}} & \multicolumn{3}{c|}{\textbf{Split CIFAR-10}} & \multicolumn{3}{c}{\textbf{Split Tiny-ImageNet}} \\
& &  & Class-IL ($\uparrow$) & Task-IL ($\uparrow$) & \begin{tabular}{@{}c@{}}FLOPs Train\\ $\times 10^{15}$ ($\downarrow$) \end{tabular} &  Class-IL ($\uparrow$) & Task-IL ($\uparrow$) & \begin{tabular}{@{}c@{}}FLOPs Train\\ $\times 10^{16}$ ($\downarrow$) \end{tabular} \\
\midrule
EWC~\cite{kirkpatrick2017overcoming} & \multirow{2}{*}{0.00} & \multirow{2}{*}{--} &  19.49\scriptsize{$\pm$0.12} & 68.29\scriptsize{$\pm$3.92} & 8.3 & 7.58\scriptsize{$\pm$0.10} & 19.20\scriptsize{$\pm$0.31}  & 13.3\\
LwF~\cite{li2017learning} &  &  & 19.61\scriptsize{$\pm$0.05} & 63.29\scriptsize{$\pm$2.35} & 8.3 & 8.46\scriptsize{$\pm$0.22} & 15.85\scriptsize{$\pm$0.58} & 13.3\\
\midrule
PackNet~\cite{mallya2018packnet} & \multirow{2}{*}{0.50$^\dagger$} & \multirow{2}{*}{--} & - & 93.73\scriptsize{$\pm$0.55} & 5.0 & -- & 61.88\scriptsize{$\pm$1.01} & 7.3\\
LPS~\cite{wang2020learn} & &  & - & 94.50\scriptsize{$\pm$0.47} & 5.0 & -- & 63.37\scriptsize{$\pm$0.83} & 7.3\\
\midrule
A-GEM \cite{chaudhry2018efficient} &\multirow{5}{*}{0.00} & \multirow{5}{*}{200} & 20.04\scriptsize{$\pm$0.34} & 83.88\scriptsize{$\pm$1.49} & 11.1 & {8.07\scriptsize{$\pm$0.08}} & 22.77\scriptsize{$\pm$0.03} & 17.8\\
iCaRL \cite{rebuffi2017icarl} & &  & 49.02\scriptsize{$\pm$3.20} & 88.99\scriptsize{$\pm$2.13} & 11.1 & {7.53\scriptsize{$\pm$0.79}} & 28.19\scriptsize{$\pm$1.47} & 17.8 \\
FDR \cite{benjamin2018measuring} &  &  & 30.91\scriptsize{$\pm$2.74} & 91.01\scriptsize{$\pm$0.68}  & 13.9 & {8.70\scriptsize{$\pm$0.19}} & 40.36\scriptsize{$\pm$0.68} & 22.2\\
ER \cite{chaudhry2019tiny} & &  & 44.79\scriptsize{$\pm$1.86} & 91.19\scriptsize{$\pm$0.94} & 11.1 & {8.49\scriptsize{$\pm$0.16}} & 38.17\scriptsize{$\pm$2.00} & 17.8 \\
DER++ \cite{buzzega2020dark} & &  & 64.88\scriptsize{$\pm$1.17} & 91.92\scriptsize{$\pm$0.60} & 13.9 & {10.96\scriptsize{$\pm$1.17}} & 40.87\scriptsize{$\pm$1.16} & 22.2 \\
\midrule
 \rowcolor[gray]{.9} \methodabbr-ER$_{75}$ &  &  
                    & 46.89\scriptsize{$\pm$0.68} & 92.02\scriptsize{$\pm$0.72} & 2.0 & 8.98\scriptsize{$\pm$0.38} & 39.14\scriptsize{$\pm$0.85} & 3.2 \\
 \rowcolor[gray]{.9} \methodabbr-DER++$_{75}$ & \multirow{-2}{*}{0.75} &  & 66.30\scriptsize{$\pm$0.98} & 94.06\scriptsize{$\pm$0.45}  & 2.5 & 12.73\scriptsize{$\pm$0.40} & 42.06\scriptsize{$\pm$0.73} & 4.0 \\
 \rowcolor[gray]{.9} \methodabbr-ER$_{90}$ &  & & 45.81\scriptsize{$\pm$1.05} & 91.49\scriptsize{$\pm$0.47} & 0.9 & 8.67\scriptsize{$\pm$0.41} & 38.79\scriptsize{$\pm$0.39} & 1.4 \\
  \rowcolor[gray]{.9} \methodabbr-DER++$_{90}$ & \multirow{-2}{*}{0.90} &\multirow{-2}{*}{200} & 65.79\scriptsize{$\pm$1.33} & 93.73\scriptsize{$\pm$0.24} & 1.1 & 12.27\scriptsize{$\pm$1.06} & 41.17\scriptsize{$\pm$1.31} & 1.8\\
 \rowcolor[gray]{.9} \methodabbr-ER$_{95}$ & & & 44.59\scriptsize{$\pm$0.23} & 91.07\scriptsize{$\pm$0.64}  & 0.5 & 8.43\scriptsize{$\pm$0.09} & 38.20\scriptsize{$\pm$0.46} & 0.8\\
 \rowcolor[gray]{.9}\methodabbr-DER++$_{95}$ & \multirow{-2}{*}{0.95} & & 65.18\scriptsize{$\pm$1.25} & 92.97\scriptsize{$\pm$0.37} & 0.6 & 10.76\scriptsize{$\pm$0.62} & 40.54\scriptsize{$\pm$0.98} & 1.0\\
 \midrule
A-GEM \cite{chaudhry2018efficient} & \multirow{5}{*}{0.00} & \multirow{5}{*}{500} & 22.67\scriptsize{$\pm$0.57} & 89.48\scriptsize{$\pm$1.45} & 11.1 & {8.06\scriptsize{$\pm$0.04}} & 25.33\scriptsize{$\pm$0.49} & 17.8\\
iCaRL \cite{rebuffi2017icarl} & &  & 47.55\scriptsize{$\pm$3.95} & 88.22\scriptsize{$\pm$2.62} & 11.1& {9.38\scriptsize{$\pm$1.53}} & 31.55\scriptsize{$\pm$3.27} & 17.8\\
FDR \cite{benjamin2018measuring} & &  & 28.71\scriptsize{$\pm$3.23} & 93.29\scriptsize{$\pm$0.59} & 13.9 & {10.54\scriptsize{$\pm$0.21}} & 49.88\scriptsize{$\pm$0.71} & 22.2\\
ER \cite{chaudhry2019tiny} & &  & 57.74\scriptsize{$\pm$0.27} & 93.61\scriptsize{$\pm$0.27} & 11.1  & {9.99\scriptsize{$\pm$0.29}} & 48.64\scriptsize{$\pm$0.46} & 17.8 \\
DER++ \cite{buzzega2020dark} & &  & 72.70\scriptsize{$\pm$1.36} & 93.88\scriptsize{$\pm$0.50} & 13.9 & {19.38\scriptsize{$\pm$1.41}} & 51.91\scriptsize{$\pm$0.68} & 22.2 \\
\midrule
 \rowcolor[gray]{.9} \methodabbr-ER$_{75}$ &   & & 60.80\scriptsize{$\pm$0.22} & 93.82\scriptsize{$\pm$0.32} & 2.0 & 10.48\scriptsize{$\pm$0.29} & 50.83\scriptsize{$\pm$0.69}  & 3.2\\
 \rowcolor[gray]{.9} \methodabbr-DER++$_{75}$ & \multirow{-2}{*}{0.75} &  & 74.09\scriptsize{$\pm$0.84} & 95.19\scriptsize{$\pm$0.34} & 2.5 & 20.75\scriptsize{$\pm$0.88} & 52.19\scriptsize{$\pm$0.43} & 4.0\\
 \rowcolor[gray]{.9} \methodabbr-ER$_{90}$ & & & 59.34\scriptsize{$\pm$0.97} & 93.33\scriptsize{$\pm$0.10} & 0.9 & 10.12\scriptsize{$\pm$0.53} & 49.46\scriptsize{$\pm$1.22} & 1.4\\
  \rowcolor[gray]{.9} \methodabbr-DER++$_{90}$ & \multirow{-2}{*}{0.90} &\multirow{-2}{*}{500} & 73.42\scriptsize{$\pm$0.95} & 94.82\scriptsize{$\pm$0.23} & 1.1 & 19.62\scriptsize{$\pm$0.67} & 51.93\scriptsize{$\pm$0.36} & 1.8\\
 \rowcolor[gray]{.9} \methodabbr-ER$_{95}$ & &  & 57.75\scriptsize{$\pm$0.45} & 92.73\scriptsize{$\pm$0.34} & 0.5 & 9.91\scriptsize{$\pm$0.17} & 48.57\scriptsize{$\pm$0.50} & 0.8\\
 \rowcolor[gray]{.9} \methodabbr-DER++$_{95}$ & \multirow{-2}{*}{0.95} & & 72.14\scriptsize{$\pm$0.78} & 94.39\scriptsize{$\pm$0.15} & 0.6 & 19.01\scriptsize{$\pm$1.32} & 51.26\scriptsize{$\pm$0.78} & 1.0 \\
\bottomrule
\multicolumn{9}{l}{$^\dagger$PackNet and LPS actually have a decreased sparsity after learning every task, we use 0.50 to roughly represent the average sparsity.}
\end{tabular}
}

\end{center}
\vspace{-0.1in}
\end{table*}

\textbf{Comparison with CL methods.} Table~\ref{table:cifar_imagenet_CL} summarizes the results on Split CIFAR-10 and Tiny-ImageNet, under both class-incremental (Class-IL) and task-incremental (Task-IL) settings. From Table~\ref{table:cifar_imagenet_CL}, we can clearly tell that \methodabbr significantly improves ER and DER++, while also outperforms other CL baselines, in terms of training efficiency (measured in FLOPs). With higher sparsity ratio, \methodabbr leads to less training FLOPs. Notably, \methodabbr achieves  $23\times$ training efficiency improvement upon DER++ with a sparsity ratio of 0.95. On the other hand, our framework also improves the average accuracy of ER and DER++ consistently under all cases with a sparsity ratio of 0.75 and 0.90, and only slight performance drop when sparsity gets larger as 0.95. In particular, \methodabbr-DER++ with 0.75 sparsity ratio sets new SOTA accuracy, with all buffer sizes under both benchmarks. The outstanding performance of \methodabbr indicates that our proposed strategies successfully preserve accuracy by further mitigating catastrophic forgetting with a much sparser model. Moreover, the improvement that \methodabbr brings to two different existing CL methods shows the generalizability of \methodabbr as a unified framework, \emph{i.e.}, it has the potential to be combined with a wide array of existing methods.

We would also like to take a closer look at PackNet and LPS, which also leverage the idea of sparsity to split the model by different tasks, a different motivation from training efficiency. Firstly, they are only compatible with the Task-IL setting, since they leverage task identity at both training and test time. Moreover, the model sparsity of these methods reduces with the increasing number of tasks, which still leads to much larger overall training FLOPs than that of \methodabbr. This further demonstrates the importance of keeping a sparse model without permanent expansion throughout the CL process.

\begin{table*}[t!]
\caption{Comparison with CL-adapted sparse training methods. All methods are combined with DER++ with a 500 buffer size. \methodabbr outperforms all methods in both accuracy and training efficiency, under all sparsity ratios. All three methods here can save $20\% \sim 51\%$ memory footprint, please see Appendix~\ref{app:memory} for details.}

\label{table:cifar_imagenet_sparse}
\begin{center}
\scalebox{0.73}{
\begin{tabular}{l||c|cc||cc}
\toprule 
 \multirow{3}{*}{\textbf{Method}} & \multirow{3}{*}{\textbf{Spasity}} & \multicolumn{2}{c||}{\textbf{Split CIFAR-10}} & \multicolumn{2}{c}{\textbf{Split Tiny-ImageNet}} \\
& &  \begin{tabular}{@{}c@{}} Class-IL ($\uparrow$) \end{tabular} & \begin{tabular}{@{}c@{}}FLOPs Train\\ $\times 10^{15}$ ($\downarrow$) \end{tabular} &  \begin{tabular}{@{}c@{}} Class-IL ($\uparrow$) \end{tabular} & \begin{tabular}{@{}c@{}}FLOPs Train\\ $\times 10^{16}$ ($\downarrow$) \end{tabular}    \\
\midrule
 DER++ \cite{buzzega2020dark} & 0.00 & 72.70\scriptsize{$\pm$1.36}  & 13.9 & {19.38\scriptsize{$\pm$1.41}} & 22.2 \\
 \midrule
  SNIP-DER++~\cite{lee2018snip} &  & 69.82\scriptsize{$\pm$0.72} & 1.6 & 16.13\scriptsize{$\pm$0.61} & 2.5 \\
 RigL-DER++~\cite{evci2020rigging} & \multirow{1}{*}{0.90} & 69.86\scriptsize{$\pm$0.59} & 1.6  & 18.36\scriptsize{$\pm$0.49} & 2.5 \\
 \rowcolor[gray]{.9} \methodabbr-DER++$_{90}$ & & 73.42\scriptsize{$\pm$0.95}& 1.1 & 19.62\scriptsize{$\pm$0.67} & 1.8 \\
 \midrule
 SNIP-DER++~\cite{lee2018snip} & & 66.07\scriptsize{$\pm$0.91} & 0.9 & 14.76\scriptsize{$\pm$0.52}  & 1.5 \\
 RigL-DER++~\cite{evci2020rigging} & \multirow{1}{*}{0.95} & 66.53\scriptsize{$\pm$1.13} & 0.9 &  15.88\scriptsize{$\pm$0.63}  & 1.5 \\
 \rowcolor[gray]{.9} \methodabbr-DER++$_{95}$ &  & 72.14\scriptsize{$\pm$0.78}& 0.6 & 19.01\scriptsize{$\pm$1.32} & 1.0 \\
\bottomrule
\end{tabular}
}
\end{center}
\vspace{-0.5cm}
\end{table*}

\textbf{Comparison with CL-adapted sparse training methods.} Table~\ref{table:cifar_imagenet_sparse} shows the result under the more difficult Class-IL setting. \methodabbr outperforms all CL-adapted sparse training methods in both accuracy and training FLOPs. The performance gap between \methodabbr-DER++ and other methods gets larger with a higher sparsity. SNIP- and RigL-DER++ achieve training acceleration at the cost of compromised accuracy, which suggests that keeping accuracy is a non-trivial challenge for existing sparse training methods under the CL setting. 
SNIP generates the static initial mask after network initialization which does not consider the structure suitability among tasks. Though RigL adopts a dynamic mask, the lack of task-aware strategy prevents it from generalizing well to the CL setting. 

\subsection{Effectiveness of Key Components} \label{sec:ablation}

\begin{table}[t]
\centering 

\captionof{table}{Ablation study on Split-CIFAR10 with 0.75 sparsity ratio. All components contributes to the overall performance, in terms of both accuracy and efficiency (training FLOPs and memory footprint).}
        \vspace{0.2cm}
        \centering 
        \scalebox{0.73}{
        \begin{tabular}{ccc|ccc}
        \toprule 
         \textbf{\sparseweightabbr}  & \textbf{\sparsedataabbr} & \textbf{\sparsegradabbr} & Class-IL ($\uparrow$)& \begin{tabular}{@{}c@{}}FLOPs Train\\ $\times 10^{15}$ ($\downarrow$) \end{tabular} & \begin{tabular}{@{}c@{}}Memory \\ Footprint ($\downarrow$) \end{tabular}   \\
        \midrule
        \xmark& \xmark & \xmark & 72.70  & 13.9 & 247MB \\
        \cmark & \xmark & \xmark & 73.37 & 3.6 & 180MB\\
        \cmark & \cmark & \xmark & 73.80 & 2.8 & 180MB\\
        \cmark& \xmark & \cmark & 73.97 &  3.3 & 177MB\\
        \cmark & \cmark & \cmark & \bf 74.09  & \bf 2.5 & \bf 177MB\\
        \bottomrule
        \end{tabular}
        }
        \label{table:ablation}
        \vspace{-0.4cm}
\end{table}

\textbf{Ablation study.} We provide a comprehensive ablation study in Table~\ref{table:ablation} using \methodabbr-DER++ with 0.75 sparsity on Split CIFAR10. Table~\ref{table:ablation} demonstrates that all components of our method contribute to both efficiency and accuracy improvements. Comparing row 1 and 2, we can see that the majority of FLOPs decrease results from \sparseweightabbr. Interestingly, \sparseweightabbr leads to an increase in accuracy, indicating \sparseweightabbr generates a sparse model that is even more suitable for learning all tasks than then full dense model. Comparing row 2 and 3, we can see that \sparsedataabbr indeed further accelerates training by removing less informative examples. As discussed in Section~\ref{sec:DDR}, when we remove a certain number of samples (30\% here), we achieve a point where we keep as much informative samples as we need, and also balance the current and buffered data. Comparing row 2 and 4, \sparsegradabbr reduce both training FLOPs and memory footprint while improve the performance of the network. Finally, the last row demonstrates the collaborative performance of all components. We also show the same ablation study with 0.90 sparsity in Appendix~\ref{app:ablation90} for reference. Detail can be found in Appendix~\ref{app:metrics}.

\begin{figure}[t]
\centering
\begin{minipage}{0.45\textwidth}
  \centering
  \includegraphics[width=.9\textwidth]{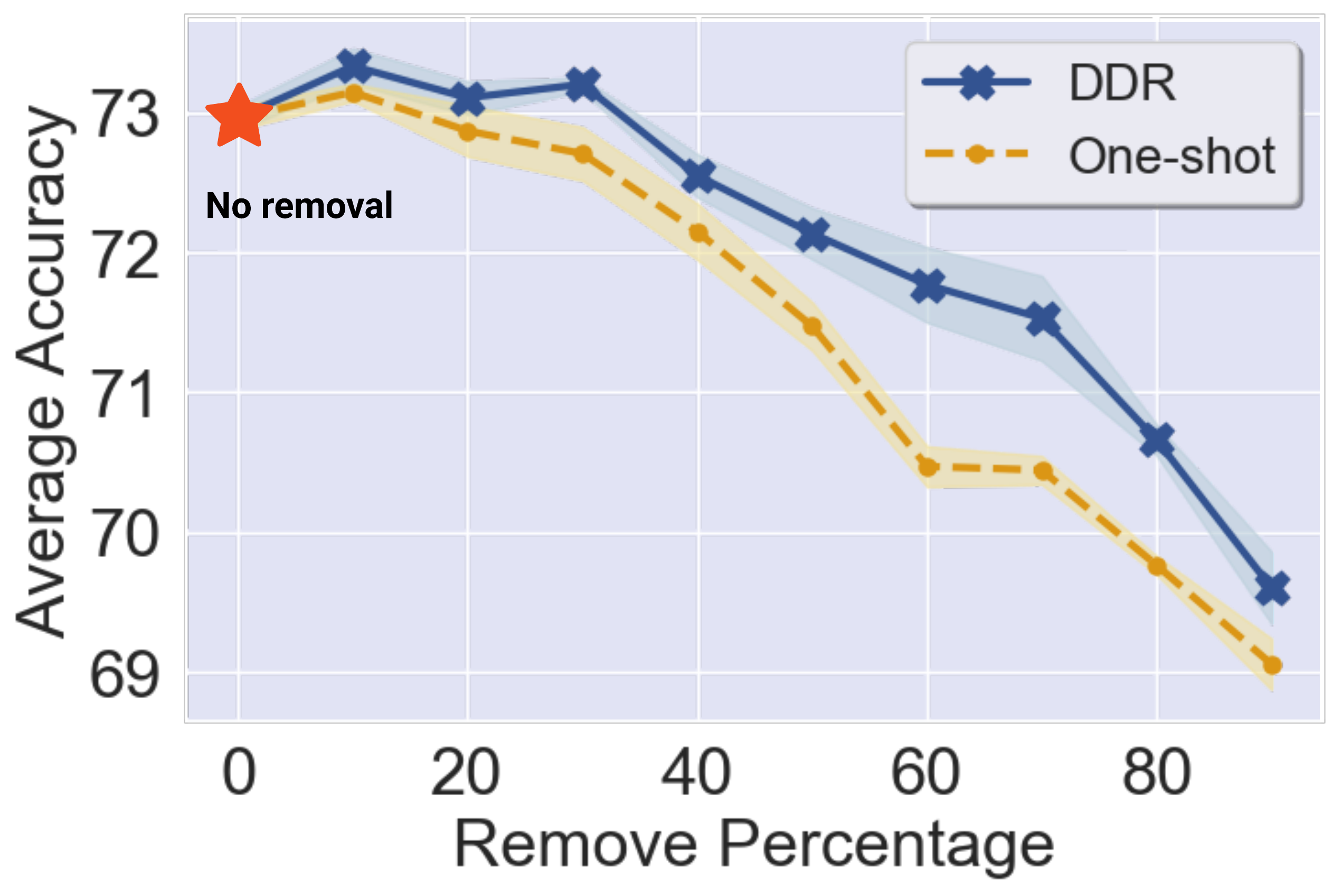}
  \captionof{figure}{Comparison between DDR and One-shot~\cite{yuan2021mest} data removal strategy w.r.t. different data removal proportion $\rho$. DDR outperforms One-shot  and also achieves better accuracy when $\rho \leq 30\%$.}
  \label{fig:de_remove_n}
\end{minipage}
\hfill
\begin{minipage}{0.42\textwidth}
\centering
  \includegraphics[width=.9\textwidth]{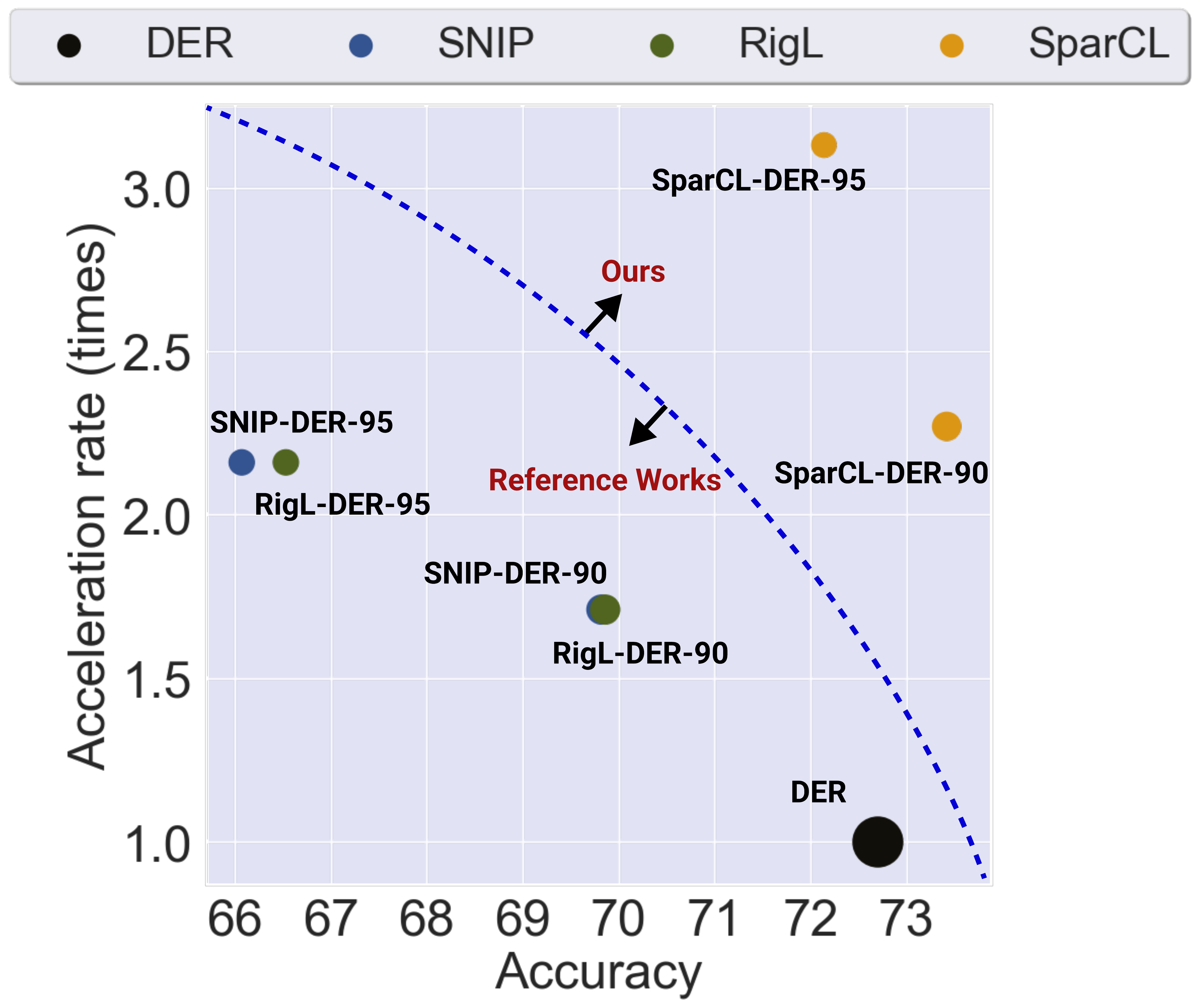}

  \captionof{figure}{Comparison with CL-adapted sparse training methods in training acceleration rate and accuracy results. The radius of circles are measured by memory footprint.}
  \label{fig:mobile}
\end{minipage}
\vspace{-0.5cm}
\end{figure}

\textbf{Exploration on \sparsedataabbr.} To understand the influence of the data removal proportion $\rho$, and the \texttt{cutoff} stage for each task, we show corresponding experiment results in Figure~\ref{fig:de_remove_n} and Appendix~\ref{app:cutoff}, respectively. In Figure~\ref{fig:de_remove_n}, we fix $\texttt{cutoff}=4$, \emph{i.e.}, gradually removing equal number of examples every 5 epochs until epoch 20, and vary $\rho$ from $10\%$ to $90\%$. We also compare DDR with One-shot removal strategy~\cite{yuan2021mest}, which removes all examples at once at \texttt{cutoff}. DDR outperforms One-shot consistently with different $\rho$ in average accuracy. Also note that since DDR removes the examples gradually before the \texttt{cutoff} stage, DDR is more efficient than One-shot. When $\rho \leq 30\%$, we also observe increased accuracy of DDR compared with the baseline without removing any data. When $\rho \geq 40\%$, the accuracy gets increasingly lower for both strategies. The intuition is that when DDR removes a proper amount of data, it removes redundant information while keeps the most informative examples. Moreover, as discussed in Section~\ref{sec:DDR}, it balances the current and buffered data, while also leave informative samples in the buffer. When DDR removes too much data, it will also lose informative examples, thus the model has not learned these examples well before removal.

\textbf{Exploration on \sparsegradabbr.} We test the efficacy of \sparsegradabbr at different sparsity levels. Detailed exploratory experiments are shown in Appendix~\ref{app:dgm} for reference. The results indicate that by setting the proportion $q$ within an appropriate range, \sparsegradabbr can consistently improve the accuracy performance regardless of the change of weight sparsity.

\subsection{Mobile Device Results} \label{sec:mobile_device}


The training acceleration results are measured on the CPU of an off-the-shelf Samsung Galaxy S20 smartphone, which has the Qualcomm Snapdragon 865 mobile platform with a Qualcomm Kryo 585 Octa-core CPU. We run each test on a batch of 32 images to denote the training speed. The detail of on-mobile compiler-level optimizations for training acceleration can be found in Appendix~\ref{app:compiler}.

The acceleration results are shown in Figure~\ref{fig:mobile}.  \methodabbr can achieve approximately 3.1$\times$ and 2.3$\times$ training acceleration with 0.95 sparsity and 0.90 sparsity, respectively. 
Besides, our framework can also save 51\% and 48\% memory footprint when the sparsity is 0.95 and 0.90.  Furthermore, the obtained sparse models save the storage consumption by using compressed sparse row (CSR) storage and can be accelerated to speed up the inference on-the-edge. We provide on-mobile inference acceleration results in Appendix~\ref{app:mobile_results}.
 

\section{Conclusion}\label{sec:Conclusion}
This paper presents a unified framework named \methodabbr for efficient CL that achieves both learning acceleration and accuracy preservation.
It comprises three complementary strategies: task-aware dynamic masking for weight sparsity, dynamic data removal for data efficiency, and dynamic gradient masking for gradient sparsity. 
Extensive experiments on standard CL benchmarks and real-world edge device evaluations demonstrate that our method significantly improves upon existing CL methods and outperforms CL-adapted sparse training methods. We discuss the limitations and potential negative social impacts of our method in Appendix~\ref{app:limitation} and~\ref{app:social}, respectively.

\section*{Checklist}


\begin{enumerate}

\item For all authors...
\begin{enumerate}
  \item Do the main claims made in the abstract and introduction accurately reflect the paper's contributions and scope?
        \answerYes{The claims match the experimental results and it is expected to generalize according to the diverse experiments stated in our paper. We include all of our code, data, and models in the supplementary materials, which can reproduce our experimental results.}
  \item Did you describe the limitations of your work?
    \answerYes{See Section~\ref{sec:Conclusion} and Appendix~\ref{app:limitation}.}
  \item Did you discuss any potential negative societal impacts of your work?
    \answerYes{See Section~\ref{sec:Conclusion} and Appendix~\ref{app:social}.}
  \item Have you read the ethics review guidelines and ensured that your paper conforms to them?
        \answerYes{We have read the ethics review guidelines and ensured that our paper conforms to them.}
\end{enumerate}

\item If you are including theoretical results...
\begin{enumerate}
  \item Did you state the full set of assumptions of all theoretical results?
    \answerNA{Our paper is based on the experimental results and we do not have any theoretical results.}
        \item Did you include complete proofs of all theoretical results?
    \answerNA{Our paper is based on the experimental results and we do not have any theoretical results.}
\end{enumerate}

\item If you ran experiments...
\begin{enumerate}
  \item Did you include the code, data, and instructions needed to reproduce the main experimental results (either in the supplemental material or as a URL)?
    \answerYes{See Section~\ref{sec:exp_setting}, Section~\ref{sec:mobile_device} and we provide code to reproduce the main experimental results.}
  \item Did you specify all the training details (e.g., data splits, hyperparameters, how they were chosen)?
    \answerYes{See Section~\ref{sec:exp_setting} and Section~\ref{sec:mobile_device}.}
        \item Did you report error bars (e.g., with respect to the random seed after running experiments multiple times)?
    \answerYes{See Table~\ref{table:cifar_imagenet_CL}, Table~\ref{table:cifar_imagenet_sparse}, fig~\ref{fig:intro}, fig~\ref{fig:de_remove_n}.}
        \item Did you include the total amount of compute and the type of resources used (e.g., type of GPUs, internal cluster, or cloud provider)?
    \answerYes{See Section~\ref{sec:exp_setting}, Section~\ref{sec:mobile_device}.}
\end{enumerate}

\item If you are using existing assets (e.g., code, data, models) or curating/releasing new assets...
\begin{enumerate}
  \item If your work uses existing assets, did you cite the creators?
    \answerYes{We mentioned and cited the datasets (Split CIFAR-10 and Tiny-ImageNet), and all comparing methods with their paper and github in it.}
  \item Did you mention the license of the assets?
    \answerYes{The licences of used datasets/models are provided in the cited references and we state them explicitly in Appendix~\ref{app:license}.}
  \item Did you include any new assets either in the supplemental material or as a URL? \answerYes{We provide code for our proposed method in the supplement.}
  \item Did you discuss whether and how consent was obtained from people whose data you're using/curating?
   \answerNA{}
  \item Did you discuss whether the data you are using/curating contains personally identifiable information or offensive content?
   \answerNA{}
\end{enumerate}

\item If you used crowdsourcing or conducted research with human subjects...
\begin{enumerate}
  \item Did you include the full text of instructions given to participants and screenshots, if applicable?
    \answerNA{}{}
  \item Did you describe any potential participant risks, with links to Institutional Review Board (IRB) approvals, if applicable?
    \answerNA{}
  \item Did you include the estimated hourly wage paid to participants and the total amount spent on participant compensation?
    \answerNA{}
\end{enumerate}

\end{enumerate}


\newpage

\appendix
\setcounter{table}{0}
\renewcommand{\thetable}{A\arabic{table}}

\section{Limitations} \label{app:limitation}

One limitation of our method is that we assume a rehearsal buffer is available throughout the CL process. Although the assumption is widely-accepted, there are still situations that a rehearsal buffer is not allowed. However, as a framework targeting for efficiency, our work has the potential to accelerate all types of CL methods. For example, simply removing the terms related to rehearsal buffer in~\eqref{eq:cwi} and~\eqref{eq:cgi} could serve as a naive variation of our method that is compatible with other non-rehearsal methods. It is interesting to further improve \methodabbr to be more generic for all kinds of CL methods. Moreover, the benchmarks we use are limited to vision domain. Although using vision-based benchmarks has been a common practice in the CL community, we believe evaluating our method, as well as other CL methods, on datasets from other domains such as NLP will lead to a more comprehensive and reliable conclusion. We will keep track of newer CL benchmarks from different domains and further improve our work correspondingly.

\section{Potential Negative Societal Impact} \label{app:social}

Although \methodabbr is a general framework to enhance efficiency for various CL methods, we still need to be aware of its potential negative societal impact. For example, we need to be very careful about the trade-off between accuracy and efficiency when using \methodabbr. If one would like to pursue efficiency by setting the sparsity ratio too high, then even \methodabbr will result in significant accuracy drop, since the over-sparsified model does not have enough representation power. Thus, we should pay much attention when applying \methodabbr on accuracy-sensitive applications such as healthcare~\cite{yu2018artificial}. Another example is that, \methodabbr as a powerful tool to make CL methods efficient, can also strengthen models for malicious applications~\cite{brundage2018malicious}. Therefore, we encourage the community to come up with more strategies and regulations to prevent malicious use of artificial intelligence.

\section{Dataset Licensing Information} \label{app:license}

\begin{itemize}
    \item CIFAR-10~\cite{krizhevsky2009learning} is licensed under the MIT license.
    \item The licensing information of Tiny-ImageNet \cite{lecun1998mnist} is not available. However, the data is available for free to researchers for non-commercial use.
\end{itemize}

\section{Additional Experiment Details and Results} \label{app:expri_overall}

We set $\alpha = 0.5, \beta = 1$ in~\eqref{eq:cwi} and~\eqref{eq:cgi}. We also set $\delta k = 5$, $p_{\texttt{inter}} = 0.01$, $p_{\texttt{intra}} = 0.005$.
We also match different weight sparsity with gradient sparsity for best performance. We sample 20\% data from Split CIFAR-10 training set for validation, and we use grid-search on this validation set to help us select the mentioned best hyperparameters. We use the same set of hyperparameters for both datasets. For accurate evaluation, we repeat each experiments 3 times using different random seeds and report the average performance. During our experiments, we adopt unstructured sparsity type and uniform sparsity ratio ($0.75, 0.90, 0.95$) for all convolutional layers in the models.

\subsection{Evaluation Metrics Explanation} \label{app:metrics}

\textbf{Training FLOPs}
The FLOPs of a single forward pass is calculated by taking the sum of the number of multiplications and additions in each layer $l$ for a given layer sparsity $s_l$. Each iteration in the training process is composed of two phases, i.e., the forward propagation and backward propagation. 

The goal of the forward pass is to calculate the loss of the current set of parameters on a given batch of data. It can be formulated as $a_l = \sigma(z_l) = \sigma(w_l * a_{l-1}+ b_l)$ for each layer $l$ in the model. Here, $w$, $b$, and $z$ represent the weights, biases, and output before activation, respectively; $\sigma(.)$
denotes the activation function; $a$ is the activations; $*$  means convolution operation. The formulation indicates that the layer activations are calculated in sequence
using the previous activations and the parameters of the layer. Activation of layers are stored in memory for the backward pass.

As for the backward propogation, the objective is to back-propagate the error signal while calculating the gradients of the parameters. The two main calculation steps can be represented as:
\begin{align}
   & \delta_l = \delta_{l+1} * \textrm{rotate}180\degree (w_l) \odot \sigma'(z_l), \label{eq:back_1}\\ 
    & G_l = a_{l-1} * \delta_l, \label{eq:back_2}
\end{align}
where $\delta_l$ is the error associated with the layer $l$, $G_l$ denotes the gradients, $\odot$ represents Hadamard product, $\sigma'(.)$ denotes the derivative of activation, and $\textrm{rotate}180\degree(.)$ means rotating the matrix by $180\degree$ is the matrix transpose operation. During the backward pass, each layer $l$ calculates two quantities, i.e., the gradient of the activations of the previous layer and the gradient of its parameters. Thus, the backward passes are counted as \textbf{twice} the computation expenses of the forward pass~\cite{evci2020rigging}.  We omit the FLOPs needed for batch normalization and cross entropy. In our work, the total FLOPs introduced by TDM, DDR, and DGM on split CIFAR-10 is approximately $4.5\times 10^{9}$ which is less than $0.0001\%$ of total training FLOPs. For split Tiny-ImageNet, the total FLOPs of them is approximately $1.8\times 10^{10}$, which is also less than $0.0001\%$ of total training FLOPs. Therefore, the computation introduced by TDM, DDR, and DGM is negligible.

\textbf{Memory Footprints} Following works ~\cite{canziani2016analysis, yuan2021mest}, the definition of memory footprints contain two parts: 1) activations (feature map pixels) during training phase, and 2) model parameters during training phase. For experiments, activations, model weights, and gradients are stored in 32-bit floating-point format for training. The memory footprint results are calculated with an approximate summation of them. 

\subsection{Details of Memory Footprint}  \label{app:memory}

The memory footprint is composed of three parts: activations, model weights, and gradients. They are all represented as $b_w$-bit numbers for training.

The number of activations in the model is the sum of the activations in each layer. Suppose that the output feature of the $l$-th layer with a batch size of $B$ is represented as $a_l \in \mathcal{R}^{B \times O_l \times H_l \times W_l}$, where $O_l$ is the number of channels and $H_l \times W_l$ is the feature size. The total number of activations of the model is thus $B \sum_l O_l H_l W_l$.

As for the model weights, our \methodabbr training a sparse model with a sparsity ratio $s\in[0,1]$ from scratch. The sparse model is obtained from a dense model with a total of $N$ weights. A higher value of $s$ indicates fewer non-zero weights in the sparse model.  Compressed sparse row (CSR) format is
commonly used for sparse storage, which greatly reduces the number of indices need to be stored for sparse matrices. As our \methodabbr adopt only one sparsity type and we use a low-bit format to store the indices, we omit the indices storage here. Therefore, the memory footprint for model representation is $(1-s)Nb_w$. 

Similar calculations can be applied for the gradient matrix. Besides the sparsity ratio $s$, additional $q$ gradients are masked out from the gradient matrix, resulting a sparsity ratio $s+q$. Therefore, the storage of gradients can be approximated as $(1-(s+q))N b_w $.

Combining the activations, model representation, and gradients, the total memory footprint
in SparCL can be represented as $(2B\sum_l O_l H_l W_l + (1-s)N + (1-(s+q))N) b_w$.

DDR requires store indices for the easier examples during the training process. The number of training examples for Split CIFAR-10 and Split Tiny-ImageNet on each task is 10000. In our work, we only need about 3KB (remove $30\%$ training data) for indices storage (in the int8 format) and the memory cost is negligible compared with the total memory footprint.

\subsection{Effect of Cutoff Stage} \label{app:cutoff}

\begin{table}[h] 
\centering
\captionof{table}{Effect of \texttt{cutoff}.}
\scalebox{0.8}{
\begin{tabular}{c|ccccccccc}
    \toprule
    \texttt{cutoff} & 1 & 2 & 3 & 4 & 5 & 6 & 7 & 8 & 9 \\
    \midrule
     Class-IL ($\uparrow$) & 71.54 & 72.38 & 72.74 & 73.20 & 73.10 & 73.32 & 73.27 & 73.08 & 73.23 \\
     \bottomrule
\end{tabular}
}
\label{app:table:cutoff}
\end{table}

To evaluate the effect of the \texttt{cutoff} stage, we use the same setting as in Figure~\ref{fig:de_remove_n} by setting the sparsity ratio to $0.90$. We keep the data removal proportion $\rho = 30\%$, and only change \texttt{cutoff}. Table~\ref{app:table:cutoff} shows the relationship between \texttt{cutoff} and the Class-IL average accuracy. Note that from the perspective of efficiency, we would like the \texttt{cutoff} stage as early as possible, so that the remaining epochs will have less examples. However, from Table~\ref{app:table:cutoff}, we can see that if we set it too early, \emph{i.e.}, $\texttt{cutoff} \leq 3$, the accuracy drop is significant. This indicate that even for the ``easy-to-learn'' examples, removing them too early results in underfitting. As a balance point between accuracy and efficiency, we choose $\texttt{cutoff} = 4$ in our final version.

\subsection{Supplementary Ablation Study} \label{app:ablation90}

\begin{table}[h]
\centering 
\captionof{table}{Ablation study on Split-CIFAR10 with 0.90 sparsity.}
        \vspace{0.2cm}
        \centering 
        \scalebox{0.8}{
        \begin{tabular}{ccc|ccc}
        \toprule 
         \textbf{\sparseweightabbr}  & \textbf{\sparsedataabbr} & \textbf{\sparsegradabbr} & Class-IL ($\uparrow$)& \begin{tabular}{@{}c@{}}FLOPs Train\\ $\times 10^{15}$ ($\downarrow$) \end{tabular} & Memory Footprint ($\downarrow$)   \\
        \midrule
        \xmark& \xmark & \xmark & 72.70  & 13.9 & 247MB \\
        \cmark & \xmark & \xmark & 72.98 & 1.6 & 166MB\\
        \cmark & \cmark & \xmark & 73.20 & 1.2 & 166MB\\
        \cmark& \xmark & \cmark & 73.30 &  1.5 & 165MB\\
        \cmark & \cmark & \cmark & \bf 73.42  & \bf 1.1 & \bf 165MB\\
        \bottomrule
        \end{tabular}
        }
        \label{app:table:ablation90}
\end{table}

Similar to Table~\ref{table:ablation}, we show ablation study with $0.90$ sparsity ratio in Table~\ref{app:table:ablation90}. Under a larger sparsity ratio, the conclusion that all components contribute to the final performance still holds. However, we can observe that the accuracy increase that comes from \sparsedataabbr and \sparsegradabbr is less than what we show in Table~\ref{table:ablation}. We assume that larger sparsity ratio makes it more difficult for the model to retain good accuracy in CL. Similar results has also been observed in~\cite{yuan2021mest} under the usual i.i.d. learning setting.

\subsection{Exploration on \sparsegradabbr} \label{app:dgm}

\begin{table}[h]
\centering 
\captionof{table}{Ablation study of the gradient sparsity ratio on Split-CIFAR10.}
        \vspace{0.2cm}
        \centering 
        \scalebox{0.8}{
        \begin{tabular}{cc|ccc}
        \toprule 
         \textbf{weight sparsity}  & {\textbf{gradient sparsity}} & Class-IL ($\uparrow$)& \begin{tabular}{@{}c@{}}FLOPs Train\\ $\times 10^{15}$ ($\downarrow$) \end{tabular}  & Memory Footprint ($\downarrow$)   \\
        \midrule
        0.75 & 0.78 & 74.08 & 3.4 & 178MB \\
        0.75 & 0.80 & 73.97 & 3.3 & 177MB\\
        0.75 & 0.82 & 73.79 & 3.3 & 177MB\\
        0.75 & 0.84 & 73.26 & 3.2 & 176MB\\
        \midrule
        0.90 & 0.91 & 73.33 & 1.6 & 166MB \\
        0.90 & 0.92 & 73.30 & 1.5 & 165MB \\
        0.90 & 0.93 & 72.64 & 1.5 & 165MB \\
        \bottomrule
        \end{tabular}
        }
        \label{table:ablation_gradient}
\end{table}

We conduct further experiments to demonstrate the influence of gradient sparsity, and the results are shown in Table~\ref{table:ablation_gradient}. There are two sets of the experiments with different weight sparsity settings: 0.75 and 0.90. Within each set of the experiments (the weight sparsity is fixed), we vary the gradient sparsity values. From the results we can see that increasing the gradient sparsity can decrease the FLOPs and memory footprint. However, the accuracy performance degrades more obvious when the gradient sparsity is too much for the weight sparsity. The results indicate that suitable gradient sparsity setting can bring further efficiency to the training process while boosting the accuracy performance. In the main results, the gradient sparsity is set as 0.80 for 0.75 weight sparsity, and set as 0.92 for 0.90 weight sparsity.  

\section{On-Mobile Compiler Optimizations and Inference Results}
\subsection{Compiler Optimizations }
\label{app:compiler}

Each iteration in the training process is composed of two phases, i.e., the forward propagation and backward propagation. Prior works \cite{he2018soft,li2019compressing,he2019filter,ma2020pconv,dong2020rtmobile,jian2021radio,gong2020privacy,zhan2021achieving} have proved that sparse weight matrices (tensors) can provide inference acceleration via reducing the number of multiplications in convolution operation. Therefore, the forward propagation phase, which is the same as inference, can be accelerated by the sparsity inherently. As for backward pass, both of the calculation steps are based on convolution, i.e.,
matrix multiplication. Equation \ref{eq:back_1} uses sparse weight matrix (tensor) as the operand, thus
can be accelerated in the same way as the forward propagation. Equation \ref{eq:back_2}  allows a sparse output
result since the gradient matrix is also sparse. Thus, both two steps have
reduced computations, which are roughly proportional to the sparsity ratio, providing the acceleration for the backward propagation phase.



Compiler optimizations are used to accelerate the inference in prior works \cite{niu2020patdnn,wu2022compiler,gong2022automatic}. In this work, we extend the compiler optimization techniques for accelerating the forward and backward pass during training on the edge devices.  Our compiler optimizations are general, support both sparse model training and inference accelerations on mobile platforms. The optimizations include 1) the supports for sparse models; 2) an auto-tuning process to determine the best-suited configurations of parameters for different mobile CPUs. The details of our compiler optimizations are presented as follows.

\subsubsection{Supports for Sparse Models}
Our framework supports sparse model training and inference accelerations with unstructured pruning. For the sparse (pruned) model, the framework first compacts the model storage with a compression format called Compressed Sparse Row (CSR) format, and then performs computation reordering to reduce the branches within each thread and eliminates the load imbalance among threads.

A row reordering optimization is also included to further improve the regularity of the weight matrix. After this reordering, the continuous rows with identical or similar numbers of non-zero weights are processed by multi-threads simultaneously, thus eliminating thread divergence and achieving load balance. Each thread processes more than one rows, thus eliminating branches and improving instruction-level parallelism.   
Moreover, a similar optimization flow (i.e., model compaction and computation reorder and other optimizations) is employed to support all compiler optimizations for sparsity as PatDNN~\cite{niu2020patdnn}.

\subsubsection{Auto-Tuning for Different Mobile CPUs}

During DNN sparse training and inference execution, there are many tuning parameters, e.g., matrix tiling sizes, loop unrolling factors, and data placement on memory, that influence the performance. It is hard to determine the best-suited configuration of these parameters manually. To alleviate this problem, our compiler incorporates an auto-tuning approach for sparse (pruned) models. The Genetic Algorithm is leveraged to explore the best-suited configurations automatically. It starts the parameter search process with an arbitrary number of chromosomes and explores the parallelism better. Acceleration codes for different DNN models and different mobile CPUs can be generated efficiently and quickly through this auto-tuning process.

\subsection{Inference Acceleration Results On Mobile} \label{app:mobile_results}

\begin{figure*} [h]
     \centering
     \includegraphics[width=0.95\textwidth]{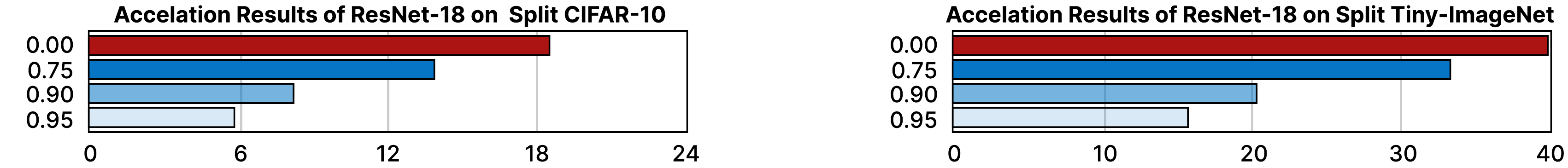}  
     \caption{Inference results of sparse models obtained from \methodabbr under different sparsity ratio compared with dense models obtained from traditional CL methods (sparsity ratio 0.00).}
    \label{fig:inference_result}  
\end{figure*}

Besides accelerating the training process, SparCL also possesses the advantages of providing a sparse model as the output for faster inference. To demonstrate this, we show the inference acceleration results of SparCL with different sparsity ratio settings on mobile in Figure \ref{fig:inference_result}. The inference time is measured on the CPU of an off-the-shelf Samsung Galaxy S20
smartphone. Each test takes 50 runs on different inputs with 8 threads on CPU. As different runs do not vary greatly, only the average time is reported. From the results we can see that the obtained sparse model from SparCL can significantly accelerate the inference on both Split-CIFAR-10 and Tiny-ImageNet dataset compared to the model obtained by traditional CL training. For ResNet-18 on Split-CIFAR-10, the model obtained by traditional CL training, which is a dense model, takes 18.53ms for inference.  The model provided by SparCL can achieve an inference time of 14.01ms, 8.30ms, and 5.85ms with sparsity ratio of 0.75, 0.90, and 0.95, respectively. The inference latency of the dense ResNet-18 obtained by traditional CL training on Tiny-ImageNet is 39.64 ms. While the sparse models provided by SparCL with sparsity ratio settings as 0.75, 0.90, and 0.95 reach inference speed of 33.06ms, 20.37ms, and 15.49ms, respectively, on Tiny-ImageNet.

\end{document}